\documentclass[runningheads]{llncs}

\usepackage{eccv}

\usepackage{eccvabbrv}

\usepackage{graphicx}
\usepackage{booktabs}

\usepackage[accsupp]{axessibility}  

\usepackage{makecell}

\usepackage{multirow}
\usepackage{indentfirst}
\usepackage{pifont}
\usepackage{amssymb}
\usepackage{amsmath}
\usepackage{amsfonts}
\usepackage{graphicx}
\usepackage[misc]{ifsym}
\usepackage{changes} 
\usepackage{colortbl}
\definechangesauthor[name={Authors}, color=red]{A.} 
\usepackage{nomencl}
\usepackage{wrapfig}

\makeatletter
\def\thanks#1{\protected@xdef\@thanks{\@thanks
        \protect\footnotetext{#1}}}
\makeatother

\usepackage{hyperref}

\usepackage{orcidlink}

\begin{document}

\title{FocusDiffuser: Perceiving Local Disparities for Camouflaged Object Detection} 

\titlerunning{FocusDiffuser}

\author{Jianwei Zhao\inst{1}\orcidlink{0009-0002-8741-7580} \and
Xin Li\inst{2}\orcidlink{0000-0001-8047-9610} \and
Fan Yang\inst{2}\orcidlink{0000-0002-1157-8719} \and
Qiang Zhai\inst{3}$^{,*}$\orcidlink{0000-0001-5328-675X}  \thanks{{*}Corresponding author: Qiang Zhai (zhaiq@sicau.edu.cn)} \and
Ao Luo\inst{4}\orcidlink{0000-0003-3494-8062}\and \\
Zicheng Jiao\inst{5}\orcidlink{0000-0002-6968-0919}\and
Hong Cheng\inst{1}\orcidlink{0000-0001-5532-9530}
}

\authorrunning{Zhao and Li et al.}


\institute{University of Electronic Science and Technology of China \and
Group 42 (G42) \and
Sichuan Agricultural University \and
Southwest Jiaotong University \and
Brown University\\
}

\maketitle

\begin{abstract}

Detecting objects seamlessly blended into their surroundings represents a complex task for both human cognitive capabilities and advanced artificial intelligence algorithms. Currently, the majority of methodologies for detecting camouflaged objects mainly focus on utilizing discriminative models with various unique designs. However, it has been observed that generative models, such as Stable Diffusion, possess stronger capabilities for understanding various objects in complex environments; Yet their potential for the cognition and detection of camouflaged objects has not been extensively explored. In this study, we present a novel denoising diffusion model, namely \emph{FocusDiffuser}, to investigate how generative models can enhance the detection and interpretation of camouflaged objects. We believe that the secret to spotting camouflaged objects lies in catching the subtle nuances in details. Consequently, our \emph{FocusDiffuser} innovatively integrates specialized enhancements, notably the Boundary-Driven LookUp (BDLU) module and Cyclic Positioning (CP) module, to elevate standard diffusion models, significantly boosting the detail-oriented analytical capabilities. Our experiments demonstrate that \emph{FocusDiffuser}, from a generative perspective, effectively addresses the challenge of camouflaged object detection, surpassing leading models on benchmarks like CAMO, COD10K and NC4K. Code and pre-trained models are available at \url{https://github.com/JWZhao-uestc/FocusDiffuser}.

\end{abstract}

\section{Introduction}
\label{sec:intro}
Camouflaged Object Detection (COD) is an important field within computer vision aimed at detecting objects that are intricately blended with their environments, rendering them nearly invisible. This technology is vital across various applications, including surveillance, search and rescue operations, and the analysis of medical imagery~\cite{liu2013model,fan2020pranet,singh2013survey, thayer1918concealing}, where the ability to identify concealed items is crucial. At present, the foremost strategies in camouflaged object detection are centered around discriminative modeling~\cite{le2019anabranch,fan2020camouflaged}. These techniques enhance conventional segmentation neural networks by weaving in a spectrum of supportive information, thereby sharpening the model's ability to discern and detect camouflaged objects within complex scenes. 

Despite the notable achievements of the \emph{de facto} paradigm in camouflaged object detection, it is constrained by intrinsic shortcomings attributable to its discriminative nature. {Firstly}, the inherent design of discriminative models restricts their capability to grasp the fundamental probabilistic structures of data~\cite{croitoru2023diffusion, sohl2015deep}, complicating the management of intricate variations and advanced camouflage techniques. {Secondly}, their primary concentration on class differentiation, rather than a deep comprehension of data distribution, hampers their ability to generalize well to new or unseen camouflage patterns. {Thirdly}, these models are always vulnerable to noise and data irregularities, significantly undermining their precision in identifying camouflaged objects. {Not to mention}, discriminative models are notably prone to overfitting on training data, which diminishes their performance in practical, real-world situations. These limitations in discriminative models spark an intriguing question: \emph{Might the advanced representational abilities of generative models offer a more effective solution to camouflaged object detection?}

\begin{figure*}[pt]
   \centering
      \includegraphics[width=0.95\linewidth]{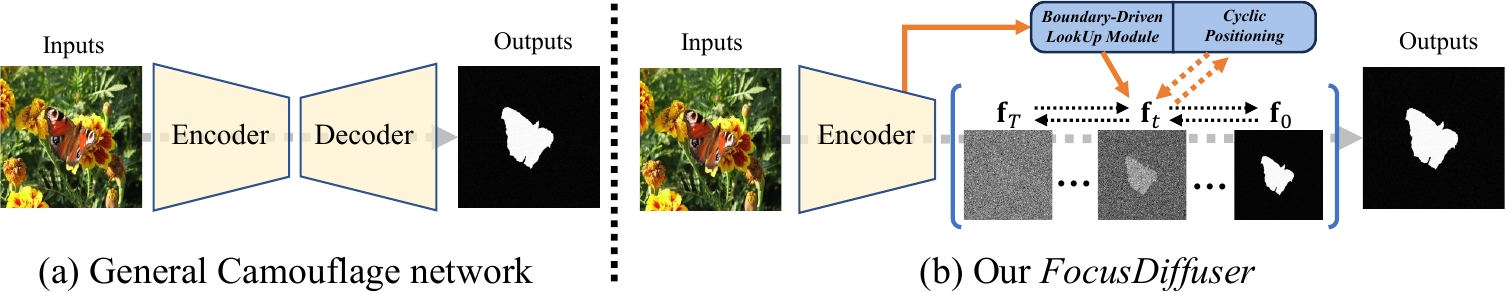}
    \vspace{-1em}
    \caption{{{\bfseries Ideal illustration}. General COD models (a) directly map images to camouflaged object maps. In contrast, \emph{FocusDiffuser} (b) generates the detection results through a reverse denoising process, based on mined information from image.} 
    \label{fig:1}}.
    \vspace{-2.5em}
\end{figure*}

With these considerations in mind, we explore generative models as a promising new direction for detecting camouflaged objects and introduce \emph{FocusDiffuser} as our proposed solution. Fig.~\ref{fig:1} illustrates the transition from discriminative mappings to a generative paradigm that effectively discerns and generates masks for camouflaged objects in scenes, guided by pertinent information as conditions. Our key concept involves detecting camouflaged objects through a `noise-to-mask' strategy, methodically filtering out noise from randomly generated patterns, guided by conditions extracted from the image itself. Specifically, during the training process, Gaussian noise is methodically integrated into the ground truth camouflaged map to synthesize an initial noise. \emph{FocusDiffuser} is adeptly calibrated to distill the noise, conditioned on meaningful information unearthed by built-in modules, thereby fostering the capability to infer the masks of camouflaged objects from stochastic beginnings. In the inference stage, \emph{FocusDiffuser} adeptly backtracks along the diffusion pathway, systematically uncovering the camouflaged objects as heatmaps.

Diverging from the standard denoising diffusion model, our vision for \emph{FocusDiffuser} is to specialize in understanding and detecting camouflaged objects, thus pioneering a novel model architecture equipped with specialized integrated modules. We believe that the key to detecting camouflaged objects lies in the details, and therefore, it is essential to amplify the model's focus on intricate details. Thus, within our \emph{FocusDiffuser}, we have configured two pivotal modules: the Boundary-Driven LookUp (BDLU) module, tasked with explicitly identifying discrepancies in details, and the Cyclic Positioning (CP) module, which adopts a broader perspective to cyclically pinpoint camouflaged objects. Our BDLU module draws inspiration from dense matching techniques, accomplishing the analysis of local discrepancies through targeted lookup operations. Additionally, we harness edge information for enhanced supervision, further refining the module's precision in identifying subtle variations. As a complement, our CP module aligns with the denoising process of diffusion, continuously pinpointing targets from a broader scope, thus enhancing the overall detection capability within the diffusion framework. As a result, our \emph{FocusDiffuser} boasts superior abilities in comprehending and identifying camouflaged objects, setting it apart from existing diffusion models across diverse application areas. Overall, our {\bf contributions} include:
\begin{itemize}
	\item{\bf  A novel generative approach for camouflaged object detection.} \textit{FocusDiffuser} represents one of the first forays into employing generative models to address the challenge of detecting camouflaged objects. We contend that this work provides a fresh viewpoint in the domain and lays the groundwork for future investigations.
 
	\item{\bf A new diffusion model to enhance detail comprehension.}  \textit{FocusDiffuser} introduces advanced capabilities for understanding fine details within the diffusion models. The integration of our BDLU and CP modules significantly augments the detail discernment capacity of diffusion models, pioneering their application in the field of camouflaged object detection.
  
	\item{\bf State-of-the-art results on widely-used benchmarks.} Our \textit{FocusDiffuser} can accurately detect camouflaged objects, achieving state-of-the-art performance on benchmarks such as CAMO, COD10K and NC4K.
\end{itemize}



\section{Related Work}
\noindent{\bfseries Camouflaged Object Detection.} 
Camouflaged object detection (COD) identifies hidden entities within their environments and has diverse applications. Initially, handcrafted features such as color and intensity variations \cite{huerta2007improving}, 3D convexity \cite{pan2011study}, and motion boundaries \cite{hou2011detection} were used. Recently, deep learning methods \cite{fan2020camouflaged, le2019anabranch, zhai2021mutual} have taken the lead, leveraging discriminative models \cite{khan2024camofocus, cong2023frequency, xie2023frequency, wu2023object, li2023locate, huang2023feature, hu2023high} to detect objects using edge \cite{zhai2021mutual, lyu2023uedg,zhai2022mgl}, depth \cite{wang2023depth, wu2023source}, and category information \cite{fan2020camouflaged, le2019anabranch}. Emerging research is exploring generative models for COD. While innovative training strategies inspired by generative models \cite{he2023strategic, yang2021uncertainty} have been developed, they do not fully exploit generative capabilities. Our work introduces a novel diffusion model variant, pioneering the application of generative models in COD, offering new insights and advancements.

\noindent{\bfseries Diffusion Models.} 
Diffusion models, a key member of the generative model family, have recently been widely applied to various AI tasks due to their scalability, stability, and superior understanding capabilities~\cite{ho2020ddpm,dhariwal2021diffusion}. In computer vision, their effectiveness in image and video generation, segmentation, and synthesis has been well-established by numerous studies~\cite{dhariwal2021diffusion,song2019generative,wang2023learning,cheng2023layoutdiffuse,Zbinden_2023_ICCV, Wu_2023_ICCV, luo2024flowdiffuser,wolleb2022diffusion}. Our paper advances diffusion models with the novel \emph{FocusDiffuser} for camouflaged object detection. Unlike the concurrent diffusion-based COD model~\cite{chen2024camodiffusion}, our approach significantly enhances detailed analysis capabilities in this domain.



\section{Our Approach}

\subsection{Preliminaries}
\noindent{\bfseries Motivation.}
The potential of generative models like Stable Diffusion~\cite{ho2020ddpm,dhariwal2021diffusion} in comprehending objects within complex environments is well-explored. Their advanced pattern recognition and iterative refinement abilities offer a fresh perspective for revealing objects concealed within their environments. Inspired by these insights, we are motivated to explore the potential of diffusion models to enhance the field of camouflaged object detection by leveraging their powerful comprehension abilities.

\noindent{\bfseries Reformulation}. 
Conventional camouflaged object detection relies on a discriminative paradigm, classifying each pixel in an image $\mathbf{I} \in \mathbb{R}^{H \times W \times 3}$ against a camouflage map $\mathbf{f} \in \mathbb{R}^{H \times W \times 1}$. This approach uses a mapping function $\mathbb{F}_{\Phi}$, trained on labeled pairs $\{\mathbf{I}_i, \mathbf{f}_i\}_{i=1}^N$, to predict presence (`1') or absence (`0') of camouflaged objects. Diverging from this, our work introduces a generative paradigm with \emph{FocusDiffuser}, a method that employs diffusion models to generate camouflage maps $\mathbf{f}$ from input images by iteratively refining noise, $\mathbf{f}_n$, guided by information extracted from $\mathbf{I}$. This approach, concisely formulated as $\mathbf{f} = \mathbb{P}_{\Theta}(\mathbf{f}_n | \mathbf{I})$ and influenced by generative training techniques, represents a fundamental shift towards leveraging the generative potential of diffusion models for COD, promising new directions in detecting camouflaged objects.

\vspace{-1.5ex}
\subsection{FocusDiffuser}
\label{sec:FocusDiffuser}
\subsubsection{Overview.}
\begin{figure*}[pt]
   \centering
      \includegraphics[width=0.95\linewidth]{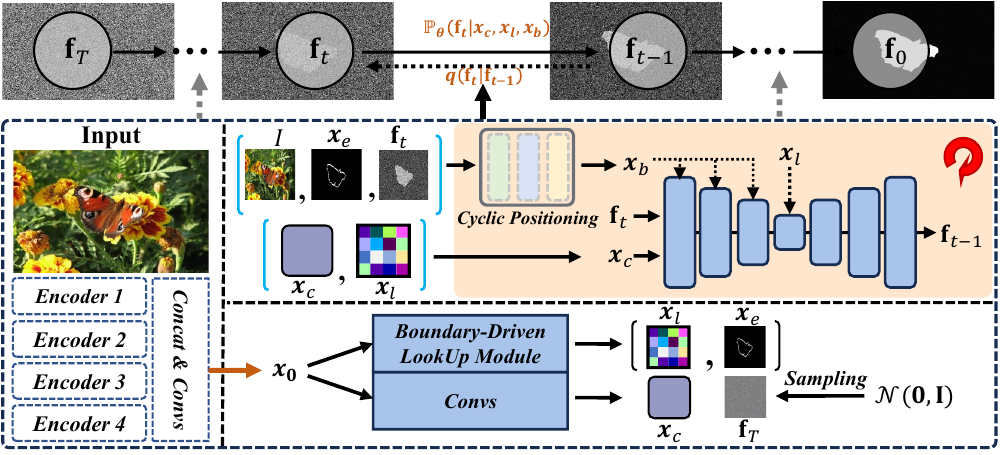}
    \vspace{-1em}
    \caption{{{\bfseries The framework of our \emph{FocusDiffuser}.} Instead of relying on the discriminative learning paradigm, our framework adopts a generative approach to guarantee reliability and generalizability. Please refer to Sec. \ref{sec:FocusDiffuser} for details.} 
    \label{fig:2}}.
    \vspace{-2.5em}
\end{figure*}

Fig.~\ref{fig:2} provides the architecture of our \emph{FocusDiffuser}, featuring an Image Conditional Encoder and a Denoising Diffusion Model with Boundary-Driven LookUp (BDLU) and Cyclic Positioning (CP) Modules. 
\begin{itemize}
\item{\bf Image Conditional Encoder.} Given an image $\mathbf{I}$, an image conditional encoder encodes it to produce hierarchical features $\{f_s\}_{s=1}^4$. These features are then combined via a convolution module to create $\boldsymbol{\mathit{x}}_0$. Subsequently, a linear embedding layer is followed to upsample $\boldsymbol{\mathit{x}}_0$, yielding $\boldsymbol{\mathit{x}}_c$. After this, $\boldsymbol{\mathit{x}}_c$ is concatenated with input noise $\mathbf{f}_t$ and introduced into the diffusion processes.

\item{\bf Boundary-Driven LookUp.} 
The Boundary-Driven LookUp (BDLU) module, receiving $\boldsymbol{\mathit{x}}_0$ as its input, produces condition feature $\boldsymbol{\mathit{x}}_l$ and edge map $\boldsymbol{\mathit{x}}_e$ as outputs. Notably, inspired by~\cite{teed2020raft,luo2023gaflow,luo2022learning}, our BDLU features a unique lookup operation designed to enhance the exploration of local detail information. This extracted information then serves as conditions to guide the diffusion process.


\item{\bf Cyclic Positioning.} At each denoising step, our Cyclic Positioning (CP) module generates a focus matrix region $M$ based on the intermediate result ${\bf{f}}_t$ from the previous iteration. This process filters out noise and directs the model's attention towards areas with camouflaged objects.

\end{itemize}
These modules combine to form our \emph{FocusDiffuser} tailored for camouflaged object detection. It boasts enhanced detail analysis capabilities and aligns with the denoising process of diffusion models to dynamically update the focus areas within a scene. Next, we delve into the details of each module.

\subsubsection{Image Conditional Encoder (ICE).}
\label{sec:Image Conditional Encoder}
Our image conditional encoder, $f_{\mathbb{ICE}}$, processes an input image $\mathbf{I} \in \mathbb{R}^{H\times W\times 3}$ to extract multi-scale hierarchical features $\{f_s\}_{s=1}^4$. These features are then passed through a convolution module to generate a high-level feature representation, $\boldsymbol{\mathit{x}}_0$. Subsequently, $\boldsymbol{\mathit{x}}_0$ is upscaled via a linear embedding layer to produce the image condition feature $\boldsymbol{\mathit{x}}_c \in \mathbb{R}^{H_n \times W_n \times d_c}$, aligning with the spatial resolution of the input noise $\mathbf{f}_t \in \mathbb{R}^{H_n \times W_n \times 1}$. This process ensures $\boldsymbol{\mathit{x}}_c$, with its $H_n \times W_n$ spatial size and $d_c$ channel depth, is optimally configured for the denoising process. 
To validate the plug-and-play capability of our \emph{FocusDiffuser}, we explore multiple variants of $f_{\mathbb{ICE}}$ : ResNet50~\cite{he2016deep}, ViT~\cite{dosovitskiy2020image}, and PVTv2~\cite{wang2022pvt}, with PVTv2 as the default choice.

\vspace{-2.5ex}
\subsubsection{Boundary-Driven LookUp (BDLU).}
\begin{figure*}[pt]
    \begin{center}
        \includegraphics[width=0.8\linewidth]{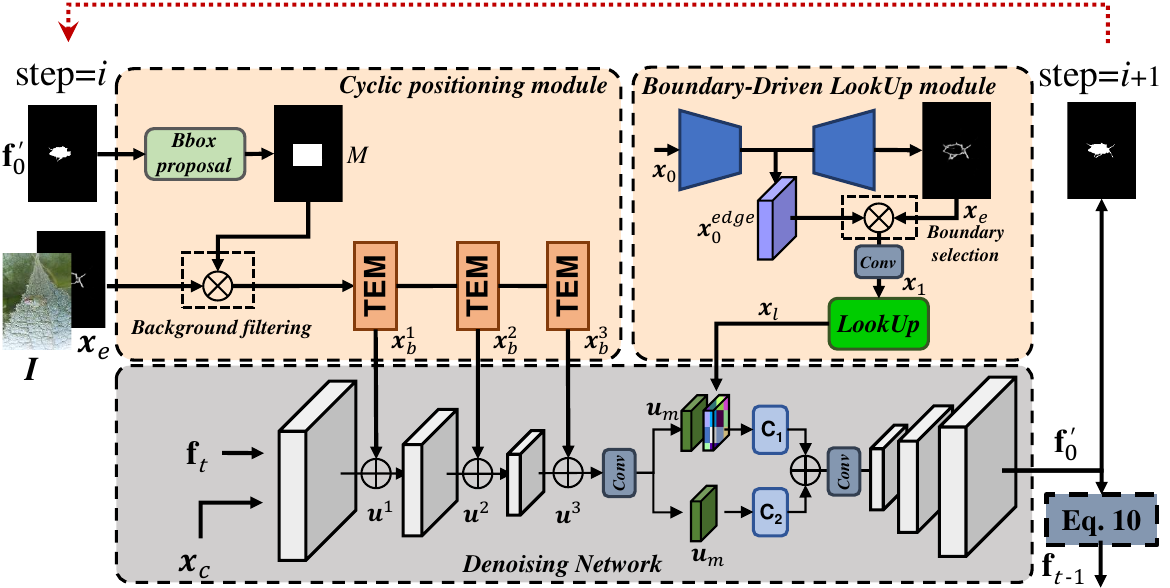}
    \end{center}
    \vspace{-1.2em}
    \caption{{\bfseries Details of Boundary-driven LookUp (BDLU) and Cyclic Positioning (CP)}. These modules are grafted onto the diffusion model and run in tandem with it. \emph{Zoom in for details}.}
    \label{fig:kao}
    \vspace{-1.5em}
\end{figure*}

Camouflaged objects integrate seamlessly into their surroundings, making the identification of their edges crucial for delineating the boundary between the object and the background. However, detecting these local disparities is challenging for global attention mechanisms due to their focus on broader contextual information. To overcome this, we incorporate the BDLU Module to maximize the utilization of local context, as illustrated in Fig.~\ref{fig:kao}. Initially, we obtain $\boldsymbol{\mathit{x}}_0$, which is then upscaled via a linear embedding layer followed by two convolution blocks to produce $\boldsymbol{\mathit{x}}_{0}^{edge} \in \mathbb{R}^{H_n \times W_n \times d_e}$. Furthermore, we introduce an additional boundary decoder composed of two convolution blocks to predict the boundary as $\boldsymbol{\mathit{x}}_e$, which is supervised using the boundary ground truth provided by the benchmarks. This process is followed by the selection of boundary-specific features, guided by $\boldsymbol{\mathit{x}}_e$. These features are then downsampled using two convolution blocks to reduce the spatial resolution, denoted as $\boldsymbol{\mathit{x}}_1 \in \mathbb{R}^{H_l \times W_l \times d_e}$, where $H_l = \frac{H_n}{8}$ and $W_l = \frac{W_n}{8}$. The downsampling process is represented as follows:
\begin{equation}
    \boldsymbol{\mathit{x}}_1 = \text{Downsample}(\boldsymbol{\mathit{x}}_{0}^{edge} \otimes \boldsymbol{\mathit{x}}_e)
\end{equation}
where $\otimes$ denotes element-wise multiplication. This method ensures a focused analysis on the vital local disparities required for effective boundary detection, optimizing the identification of camouflaged objects by leveraging the most relevant features.

It's worth emphasizing that the core of our BDLU module is an advanced lookup operation, drawing inspiration from the latest developments in dense matching technologies \cite{teed2020raft,luo2022learning2}. Yet, we've refined and adapted this concept to markedly improve the analysis of detail-rich information. This evolution ensures a more nuanced and effective methodology for detecting the intricacies of camouflaged objects, highlighting the diffusion model to enhance detail perception in complex visual environments. It primarily comprises two essential elements: \emph{Correlation Volume Pyramid} and \emph{Correlation LookUp}:

\noindent{\bfseries \small {(a)} \emph{\small Correlation Volume Pyramid:}}
Given $\boldsymbol{\mathit{x}}_1$, we initiate the process by computing visual similarity through the construction of a comprehensive correlation volume. Unlike \cite{teed2020raft}, this volume, denoted as $\mathbf{V}$, is derived from the dot product of $\boldsymbol{\mathit{x}}_1$ with itself, yielding a matrix that captures similarity across all spatial dimensions:
\begin{align}
 \mathbf{V}(\boldsymbol{\mathit{x}}_1, \boldsymbol{\mathit{x}}_1) = \mathbb{R}^{H_l \times W_l \times H_l \times W_l}
\end{align}

To discern both macroscopic and microscopic local variances, we apply pooling across the latter two dimensions of $\mathbf{V}$ using kernel sizes of 1, 2, 4, and 8, each with an equivalent stride. This operation constructs a Correlation Volume Pyramid, represented as $\{\mathbf{V}_i\}_{i=1}^{4}$, where each $\mathbf{V}_i$ inhabits a progressively reduced space $\mathbb{R}^{H_l \times W_l \times \frac{H_l}{2^i} \times \frac{W_l}{2^i}}$, efficiently capturing disparity scales.

\noindent{\bfseries (b) \emph{\small Correlation LookUp:}}
We refine the feature map by harnessing visual similarities from the correlation pyramid, denoted as $L_V$. For every pixel $x = (u, v)$ in the image $\mathbf{I}$, a local grid with radius \(\boldsymbol{r}\) outlines its neighborhood:
\begin{align}
{\mathcal{G}{(x)}}_r = \{(u+x', v+y') | (x', y') \in \mathbb{Z}^2, \|x'-y'\| \leq r \}
\end{align}

This grid, $\mathcal{G}{(x)}_r$, indexes similarity scores from $\mathbf{V}$, extracting ${(2r+1)}^2$ values per pixel to form a detailed feature map $\mathbf{L} \in \mathbb{R}^{H_l \times W_l \times {(2r+1)}^2}$. This procedure is replicated across the correlation pyramid's levels, adjusting for spatial resolutions to gather a multi-scale feature representation, $\boldsymbol{\mathit{x}}_l \in \mathbb{R}^{H_l \times W_l \times 4{(2r+1)}^2}$, offering a nuanced depiction of visual similarities.


Furthermore, the feature map $\boldsymbol{\mathit{x}}_l$ is concatenated with the latent U-net encoder feature $\boldsymbol{\mathit{u}}_m$, and fed into a convolution block, $\mathbf{C_1}$. Simultaneously, $\boldsymbol{\mathit{u}}_m$ undergoes processing by a separate convolution block, $\mathbf{C_2}$. The outputs of both blocks are then directly combined, as shown:
\begin{equation}
\begin{array}{ll}
\boldsymbol{\mathit{x}}_r & = \mathbf{C_1}([\boldsymbol{\mathit{u}}_m, \boldsymbol{\mathit{x}}_l]); \\ \boldsymbol{\mathit{u}}_m &= \mathbf{C_2}(\boldsymbol{\mathit{u}}_m) + \boldsymbol{\mathit{x}}_r
\end{array}
\end{equation}

This operation synergizes detailed feature maps with latent features for enhanced representation efficiency.

\subsubsection{Cyclic Positioning (CP):}
The loss of details from successive downsampling in conditional encoders significantly impacts the effectiveness of generative diffusion models, particularly in terms of segmentation precision. To complement the cyclical denoising steps inherent in diffusion models, we introduce the Cyclic Positioning module, an innovative approach aimed at accurately localizing camouflaged object regions with each denoising iteration.

Guided by intermediate results, we construct a bounding rectangle, indicating the predicted object from previous denoising iteration predictions, which allows us to create a binary mask ${M}_i$ for each denoising iteration $i$, ranging from 1 to $N$. This mask assigns 1 inside the rectangle and 0 outside, isolating the object of interest and eliminating redundant backgrounds. 
We leverage ${M}_i$ to isolate areas containing camouflaged objects through element-wise multiplication with the concatenated input image $\mathbf{I}$ and boundary predictions $\boldsymbol{\mathit{x}}_e$. This operation produces a new 4-channel image $\mathbf{I}_o$, which effectively removes unnecessary backgrounds, serving as enhanced detail input for the subsequent $(i+1)$-th inference cycle.
\begin{align}
\mathbf{I}_o = M \otimes [\mathbf{I}, \boldsymbol{\mathit{x}}_e]
\end{align}

It is important to highlight that initially, there is no pre-existing proposal mask for the first inference round, resulting in ${M}_0$ being set to zero. We incorporate $\mathbf{I}_o$ as an additional conditioning factor for the diffusion model, with a meticulously crafted integration process. The architecture is detailed in Fig.~\ref{fig:kao}, where to enrich texture details, a three-tier structure featuring cascaded Texture Enhance Modules (TEM)~\cite{fan2022concealed} is employed to generate a series of enhanced features $\boldsymbol{\mathit{x}}_b \in \{\boldsymbol{\mathit{x}}_{b}^{i}\}_{i=1}^{3}$. Inspired by the intricacies of the human visual system, TEM represents a highly specialized multi-branch and multi-scale convolution module. Specifically, the conditions $\{\boldsymbol{\mathit{x}}_{b}^{i}\}_{i=1}^{3}$ are directly combined with the output features $\{\boldsymbol{\mathit{u}}^{i}\}_{i=1}^{3}$ from the initial three stages of the U-net encoder integrated into the diffusion model, as described by the following formula:
\begin{align}
\boldsymbol{\mathit{u}}^{i} = \boldsymbol{\mathit{u}}^{i} + \boldsymbol{\mathit{x}}_{b}^{i}, \quad i=1,2,3
\end{align}

\begin{figure}[pt]
   \centering
      \includegraphics[width=0.98\linewidth]{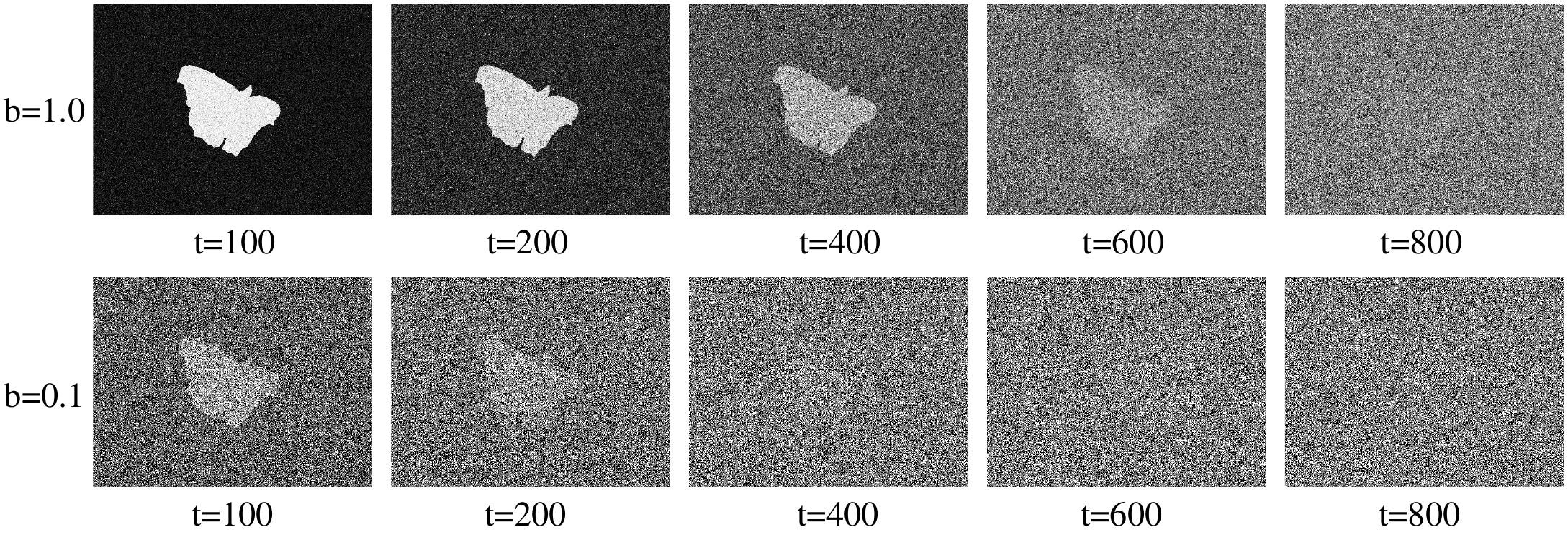}
    \vspace{-1em}
    \caption{{{\bf Visualization comparisons} of noisy masks changing over time \emph{t} under various scale factor \(\boldsymbol{b}\) settings.} \label{fig:scale factor b}}.
    \vspace{-2.5em}
\end{figure}

\subsubsection{Forward Diffusion:}
The Forward Diffusion (FD) step is a pivotal aspect of diffusion models, aimed at creating noisy masks from clean inputs by iteratively adding noise, a process vital for training. This step adheres to a Markov process, mathematically described as follows:
\begin{align}
q(\mathbf{f}_{t}|\mathbf{f}_{t-1}) = \mathcal{N}(\mathbf{f}_{t}; \sqrt{1-{\beta}_{t}}\mathbf{f}_{t-1}, {\beta}_{t}\mathbf{I})
\end{align}
Here, ${\beta}_t$ denotes the variance parameters of the Gaussian noise added at each step $t$. As $t$ progresses from 1 to $T$, $\mathbf{f}_t$ evolves from the original camouflaged map $\mathbf{f}_0$ through the process:
\begin{align}
q(\mathbf{f}_{t}|\mathbf{f}_{0}) = \mathcal{N}(\mathbf{f}_{t}; \sqrt{\bar{\alpha}_{t}}\mathbf{f}_{0}, (1-\bar{\alpha}_{t})\mathbf{I})
\end{align}
with $\bar{\alpha}_{t} = \prod_{j=0}^{t}{{\alpha}_{j}}$ and ${\alpha}_{j} = 1-{\beta}_{j}$. It’s essential to adjust $\mathbf{f}_{0}$ to fit within the range [-\(\boldsymbol{b}\), \(\boldsymbol{b}\)] before implementing FD:
\begin{align}
\mathbf{f}_{0} = (2\mathbf{f}_{0} - 1) \times \boldsymbol{b} 
\end{align}
The coefficient \(\boldsymbol{b}\) plays a crucial role in modulating the signal-to-noise ratio of the diffusion (see Fig.~\ref{fig:scale factor b}), directly impacting the denoising outcome. Sec. \ref{sec:scale-factor-b} provides a detailed examination of $b$ and its effects.

\subsubsection{Reverse Denoising:}
Starting with initial noise $\mathbf{f}_T$ from a standard Gaussian distribution, diffusion models execute a sequence of transitions from ${\mathbf{f}_T} \rightarrow \mathbf{f}_{T-\Delta } \rightarrow \dots \rightarrow \mathbf{f}_{0}$, progressively refining $\mathbf{f}_T$ towards a refined camouflaged map $\mathbf{f}_0$. This is achieved by iteratively applying a denoising function $\mathbb{P}_{\Theta}(\mathbf{f}_t | \mathbf{I})$ to gradually denoise $\mathbf{f}_T$.

Following~\cite{song2020denoising}, our approach directly estimates the predicted original camouflaged map $\mathbf{f}_0^{'} = \mathbb{P}_{\Theta}(\mathbf{f}_t | \mathbf{I})$ at each denoising execution, significantly reducing the inference steps to just 4 iterations. The reverse denoising dynamic is captured by:
\begin{align}
\mathbf{f}_{t-1} &= \sqrt{{\alpha}_{t-1}} \mathbb{P}_{\Theta}(\mathbf{f}_t | \mathbf{I}) + \sqrt{1-{\alpha}_{t-1}-{\sigma}^{2}_t} {\tilde{\epsilon}}_t + {\sigma}_t {\epsilon}_t \\
{\tilde{\epsilon}}_t &= \frac{\mathbf{f}_{t} - \sqrt{{\alpha}_{t}} \mathbb{P}_{\Theta}(\mathbf{f}_t | \mathbf{I})}{\sqrt{1-{\alpha}_{t}}}
\end{align}
Here, $\mathbb{P}_{\Theta}(\mathbf{f}_t | \mathbf{I})$ reflects the denoising prediction at time $t$, ${\tilde{\epsilon}}_t$ approximates the noise at that step, and ${\epsilon}_t$ is random Gaussian noise, illustrating a precise mechanism for transitioning from noisy to clean states within the diffusion process.

\subsubsection{Training Objective:} 
\emph{FocusDiffuser} aims to predict the original camouflaged map $\mathbf{f}_0$ directly, which necessitates measuring the loss between the conditional denoising outcome $\mathbf{f}_0^{'}$ and the actual ground truth $\mathbf{f}_0$ to guide the optimization of parameters $\Theta$. To ensure high-quality generative camouflaged maps, Mean Squared Error (MSE) loss is employed. Moreover, to accentuate the separation between the camouflaged object and its background, we integrate Binary Cross-Entropy (BCE) and Intersection over Union (IoU) loss into the training objective. The overall training objective is represented as:
\begin{align}
\mathcal{L} = \mathbf{E}_{\mathbf{f}_0^{'} \sim \mathbb{P}_{\Theta}(\mathbf{f}_t | \mathbf{I}),  t \sim [1,T], \mathbf{f}_0}\left[MSE(\mathbf{f}_0^{'} ,\mathbf{f}_{0}) + BCE(\mathbf{f}^{'}_0 ,\mathbf{f}_0) + IoU(\mathbf{f}_0^{'} ,\mathbf{f}_{0})\right]
\end{align}
This formulation encapsulates the loss calculation, employing MSE for accuracy, BCE for binary classification of foreground and background, and IoU for spatial overlap, holistically enhancing the model’s performance on camouflaged object prediction.

\section{Experiments}
\label{sec:Experiments}

\subsection{Experimental Setup}
\subsubsection{Datasets:} We evaluate our \emph{FocusDiffuser} using three renowned Camouflaged Object Detection (COD) benchmarks: CAMO~\cite{le2019anabranch}, COD10K~\cite{fan2022concealed}, and NC4K~\cite{lv2021simultaneously}. The CAMO dataset comprises 2,500 images, with an equal split between images containing camouflaged objects and those without. The COD10K dataset is more comprehensive, featuring 5,066 images with camouflaged objects, in addition to 3,000 background images and 1,934 non-camouflaged images, the latter often serving as negative controls. Consistent with prior studies~\cite{jia2022segment, fan2020camouflaged, huang2023feature}, for training, we select subsets of 1,000 camouflaged images from CAMO and 3,040 from COD10K. For validation, we set aside 250 and 2,026 camouflaged images from these datasets, respectively. Additionally, the NC4K dataset, reserved exclusively for validation, tests the \emph{FocusDiffuser}'s ability to generalize across different data scenarios.

\vspace{-2.5ex}
\subsubsection{Metrics:} To fairly evaluate \emph{FocusDiffuser}, we employ four key metrics recognized in the COD community: structure measure (\boldmath{$S_{\alpha}$}), mean E-measure ($E_{\phi}$), weighted F-measure ($F_{\beta}^{w}$), and mean absolute error ($M$). Higher values denote better performance for $S_{\alpha}$, $E_{\phi}$, and $F_{\beta}^{w}$, indicated by an upward arrow (\boldmath{$\uparrow$}), while lower values for $M$ signify improved accuracy, marked by a downward arrow (\boldmath{$\downarrow$}).


\vspace{-2.5ex}
\subsubsection{Training Settings:}
\emph{FocusDiffuser} was built using the Pytorch and trained over approximately 200 epochs with batch sizes of 8 on an NVIDIA Tesla A100 GPU, ensuring full convergence. The SGD optimizer, set with a momentum of 0.9 and an initial learning rate of 0.001, facilitated the training process. This rate was progressively reduced following a cosine schedule. To optimize computational efficiency, input images were resized to 384$\times$384 pixels, whereas the spatial noise resolution was adjusted to 288$\times$288 pixels.


\vspace{-2.5ex}

\begin{table}[pt]
    \begin{minipage}{0.43\textwidth}
    
    \caption{{\bf Detailed analysis} for BDLU and CP.}
          \resizebox{0.96\linewidth}{!}{
            \begin{tabular}{cccccccccc}
            \toprule
            \multicolumn{2}{c}{\makebox[0.15\textwidth][c]{\multirow{2}[2]{*}{Method}}} & \multicolumn{4}{c}{COD10K(2026)} & \multicolumn{4}{c}{NC4K(4121)} \\
            \cmidrule{3-10} 
            & & {$S_{\alpha}$ \boldmath{$\uparrow$}} & {$E_{\phi}$ \boldmath{$\uparrow$}} & {$F^{w}_{\beta}$ \boldmath{$\uparrow$}} & {$M$ \boldmath{$\downarrow$}} & {$S_{\alpha}$ \boldmath{$\uparrow$}} & {$E_{\phi}$ \boldmath{$\uparrow$}} & {$F^{w}_{\beta}$ \boldmath{$\uparrow$}} & {$M$ \boldmath{$\downarrow$}} \\
            \midrule
            \rowcolor{gray!40} {B} & {L} & \multicolumn{8}{c}{Boundary selection (B) \& LookUp (L) in BDLU.} \\
            \midrule
            \ding{55}&\ding{55} & {0.869} & {0.927} & {0.799} & {0.025} & {0.882} & {0.930} & {0.939} & 0.035  \\
            \ding{55}&\ding{52} & {0.867} & {0.921} & {0.794} & {0.026} & {0.877} & {0.927} & {0.936} & 0.037 \\
            \ding{52}&\ding{55} & {0.872} & {0.933} & {0.801} & {0.022} & {0.889} & {0.935} & {0.846} & 0.033  \\
            \ding{52}&\ding{52} & \textbf{0.875} & \textbf{0.939} & \textbf{0.809} & \textbf{0.020} & \textbf{0.891} & \textbf{0.940} & \textbf{0.854} & \textbf{0.029}\\
            
            \midrule
            \rowcolor{gray!40}  \multicolumn{10}{c}{Radius \(\boldsymbol{x}_1\) of grid during Correlation LookUp in BDLU.} \\ 
            \midrule
            \multicolumn{2}{c}{\emph{r}=1} & {0.870} & {0.931} & {0.800} & {0.024} & {0.885} & {0.935} & {0.847} & 0.033  \\
            \multicolumn{2}{c}{\emph{r}=2} & \textbf{0.875} & \textbf{0.939} & \textbf{0.809} & \textbf{0.020} & \textbf{0.891} & \textbf{0.940} & \textbf{0.854} & \textbf{0.029}  \\
            \multicolumn{2}{c}{\emph{r}=3} & {0.873} & {0.936} & {0.805} & {0.021} & {0.888} & {0.939} & {0.852} & 0.030  \\ 
            \multicolumn{2}{c}{\emph{r}=4} & {0.865} & {0.928} & {0.801} & {0.026} & {0.882} & {0.933} & {0.847} & 0.035  \\ 
            \midrule
            \rowcolor{gray!40} \multicolumn{10}{c}{Background filtering operation in CP guided by Mask.} \\
            \midrule
            \multicolumn{2}{c}{a.} & {0.871} & {0.934} & {0.804} & {0.023} & {0.879} & {0.934} & {0.846} & 0.034  \\
            \multicolumn{2}{c}{b.} & \textbf{0.875} & \textbf{0.939} & \textbf{0.809} & \textbf{0.020} & \textbf{0.891} & \textbf{0.940} & \textbf{0.854} & \textbf{0.029} \\  
            \multicolumn{2}{c}{c.} & {0.867} & {0.925} & {0.800} & {0.026} & {0.875} & {0.929} & {0.838} & 0.035   \\
            
            \bottomrule 
            \label{tab:BDLU-CP-detail}
           \end{tabular}%
        }
    \end{minipage}
		\quad\quad
		\begin{minipage}{0.48\textwidth}
			\caption{{\bf Ablation study} for scale factor \(\boldsymbol{b}\) and denoising step \(\boldsymbol{N}\).}
            \resizebox{0.99\linewidth}{!}{
                \begin{tabular}{cc|cccccccc}
                \toprule
                \multicolumn{2}{c}{\makebox[0.1\textwidth][c]{\multirow{2}[2]{*}{Method}}} &
                \multicolumn{4}{c}{{COD10K(2026)}} & 
                \multicolumn{4}{c}{{NC4K(4121)}} \\
                \cmidrule{3-10} 
                \multicolumn{2}{c}{}& {$S_{\alpha}$ \boldmath{$\uparrow$}} & {$E_{\phi}$ \boldmath{$\uparrow$}} & {$F^{w}_{\beta}$ \boldmath{$\uparrow$}} & {$M$ \boldmath{$\downarrow$}} & {$S_{\alpha}$ \boldmath{$\uparrow$}} & {$E_{\phi}$ \boldmath{$\uparrow$}} & {$F^{w}_{\beta}$ \boldmath{$\uparrow$}} & {$M$ \boldmath{$\downarrow$}} \\
                \midrule
                \rowcolor{gray!40} \multicolumn{10}{c}{denoising steps \(\boldsymbol{N}\).} \\
                \midrule
                {\emph{b}=0.5}& \emph{N}=2 & {0.860} & {0.925} & {0.786} & {0.027} & {0.877} & {0.926} & {0.828} & 0.038  \\
                {\emph{b}=0.5}&\emph{N}=4& \textbf{0.867} & \textbf{0.931} & \textbf{0.801} & \textbf{0.023} & \textbf{0.883} & \textbf{0.933} & \textbf{0.836} & \textbf{0.033}  \\
                {\emph{b}=0.5}&\emph{N}=6& {0.866} & \textbf{0.931} & {0.800} & {0.024} & \textbf{0.883} & {0.931} & {0.835} & \textbf{0.033} \\
                {\emph{b}=0.5}&\emph{N}=8& {0.866} & {0.930} & {0.798} & {0.024} & {0.881} & {0.929} & {0.833} & 0.034  \\
                \midrule
                {\emph{b}=0.1}&\emph{N}=2& {0.870} & {0.932} & {0.796} & {0.027} & {0.882} & {0.931} & {0.837} & 0.038  \\
                {\emph{b}=0.1}&\emph{N}=4& \textbf{0.875} & {0.939} & \textbf{0.809} & \textbf{0.020} & \textbf{0.891} & \textbf{0.940} & \textbf{0.854} & \textbf{0.029}  \\
                {\emph{b}=0.1}&\emph{N}=6& {0.874} & {0.939} & {0.808} & {0.021} & {0.890} & {0.939} & \textbf{0.854} & {0.032} \\
                {\emph{b}=0.1}&\emph{N}=8& \textbf{0.875} & \textbf{0.940} & {0.808} & \textbf{0.020} & {0.889} & {0.938} & {0.853} & \textbf{0.029}  \\
                \midrule
                \rowcolor{gray!40} \multicolumn{10}{c}{scale factor \(\boldsymbol{b}\).} \\
                \midrule
                {\emph{N}=4}& \emph{b}=1.0 & {0.859} & {0.923} & {0.774} & {0.025} & {0.875} & {0.923} & {0.813} & 0.040  \\
                {\emph{N}=4}&\emph{b}=0.5& {0.867} & {0.931} & {0.801} & {0.023} & {0.883} & {0.933} & {0.836} & {0.033}  \\
                {\emph{N}=4}&\emph{b}=0.1& \textbf{0.875} & \textbf{0.939} & \textbf{0.809} & \textbf{0.020} & \textbf{0.891} & \textbf{0.940} & \textbf{0.854} & \textbf{0.029} \\
                {\emph{N}=4}&\emph{b}=0.05& {0.870} & {0.934} & {0.800} & {0.024} & {0.886} & {0.936} & {0.850} & 0.032  \\
                \bottomrule 
                \label{tab:denoising}
                
               \end{tabular}%
            }
		\end{minipage}
        \vspace{-2em}
	\end{table}

\subsection{Main Results}
\subsubsection{Quantitative Analysis:} 
To show the effectiveness of \emph{FocusDiffuser}, we conducted a comprehensive comparison with 22 state-of-the-art (SOTA) models. This comparison was grounded in fairness, drawing results either directly from published studies or using their available trained models. As presented in Tab.~\ref{tab:main_results}, our model sets new benchmarks, leading across all metrics and datasets. Specifically, on the CAMO dataset, \emph{FocusDiffuser} led the nearest competitor by 2.3\% in \(S_{\alpha}\), 2.4\% in \(E_{\phi}\), 3.4\% in \(F_{\beta}^{w}\), and 0.8\% in \(M\), highlighting its superior detection capability in environments with complex camouflage patterns. Its proficiency was further demonstrated on the COD10K dataset, where it achieved the highest structure measure (\(S_{\alpha} = 0.875\)) and the lowest mean absolute error (\(M = 0.020\)), highlighting the model's exceptional ability to discern details and focus on concealed edge information. Remarkably, on the NC4K dataset, \emph{FocusDiffuser} exceeded the performance of existing state-of-the-art models without the need for further training, illustrating its strong generalization capabilities. This comprehensive performance not only proves \emph{FocusDiffuser}'s adeptness at identifying camouflaged objects with high precision but also sets a new standard for the field, combining nuanced detection with robust adaptability.

\begin{table}[t!]
  \centering
  \caption{{\bf Quantitative comparisons} between \emph{FocusDiffuser} and 22 SOTAs on CAMO, COD10K, and NC4K. The best results are highlighted in bold while the second-best is underlined. `-$R$', `-$V$' and  `-$P$' represent ResNet50~\cite{he2016deep}, ViT~\cite{dosovitskiy2020image} and PVTv2~\cite{wang2022pvt}, respectively. `*' represents the method based on the stronger Res2Net50~\cite{gao2019res2net}. For fair comparison, the results of ResNet50-based SOTA methods are cited under single input scale settings. The best results are bolded, and the second-best are underlined.
  }
  \resizebox{1.0\linewidth}{!}{
    \begin{tabular}{lccccccccccccc}
    \toprule
    {\multirow{2}[1]{*}{\textbf{Models}}} & {\multirow{2}[1]{*}{\textbf{Publication}}}
    & \multicolumn{4}{c}{CAMO(250)} & \multicolumn{4}{c}{COD10K(2026)} & \multicolumn{4}{c}{NC4K(4121)} \\
    \cmidrule(lr){3-14}
    \multicolumn{2}{c}{} &
    {$S_{\alpha}$ \boldmath{$\uparrow$}} & {$E_{\phi}$ \boldmath{$\uparrow$}} & {$F^{w}_{\beta}$ \boldmath{$\uparrow$}} & {$M$ \boldmath{$\downarrow$}} & 
    {$S_{\alpha}$ \boldmath{$\uparrow$}} & {$E_{\phi}$ \boldmath{$\uparrow$}} & {$F^{w}_{\beta}$ \boldmath{$\uparrow$}} & {$M$ \boldmath{$\downarrow$}} & 
    {$S_{\alpha}$ \boldmath{$\uparrow$}} & {$E_{\phi}$ \boldmath{$\uparrow$}} & {$F^{w}_{\beta}$ \boldmath{$\uparrow$}} & {$M$ \boldmath{$\downarrow$}} \\
    \midrule
    \multicolumn{14}{c}{ResNet-50 based methods.} \\
    \midrule
    
    {PFNet~\cite{mei2021camouflaged}} & {CVPR21} & {0.782} & {0.852} & {0.695} & {0.085} & {0.800} & {0.868} & {0.660} & {0.040} & {0.829} & {0.892}& {0.745} & {0.053} \\

    {UJSC~\cite{li2021uncertainty}} & {CVPR21} & {0.803} & {0.853} & {0.728} & {0.076} & {0.817} & {0.892} & {0.684} & {0.035} & {0.842} & {0.907}& {0.771} & {0.047} \\

    {MGLR~\cite{zhai2021mutual}} & {CVPR21} & {0.782} & {0.847} & {0.695} & {0.085} & {0.814} & {0.865} & {0.666} & {0.035} & {0.833} & {0.893}& {0.739} & {0.053} \\
    {UGTR~\cite{yang2021uncertainty}} & {ICCV21} & {0.785} & {0.859} & {0.794} & {0.086} & {0.818} & {0.850} & {0.667} & {0.035} & {0.839} & {‡}& {0.746} & {0.052} \\
    {PreyNet~\cite{zhang2022preynet}} & {MM22} & {0.790} & {0.842} & {0.708} & {0.077} & {0.813} & {0.881} & {0.697} & {0.034} & {‡} & {‡} & {‡} & {‡} \\
    {BSA-Net~\cite{zhu2022can}} & {AAAI22} & {0.794} & {0.851} & {0.717} & {0.079} & {0.818} & {0.891} & {0.699} & {0.034} & {0.841} & {0.897} & {0.771} & {0.048} \\
    {OCE-Net~\cite{liu2022modeling}} & {WACV22} & {0.802} & {0.852} & {0.749} & {0.073} & {0.831} & {0.901} & {0.722} & {0.033} & {0.851} & {0.907} & {0.788} & {0.044} \\
    {SegMaR~\cite{jia2022segment}} & {CVPR22} & {0.815} & {0.865} & {0.742} & {0.071} & {0.833} & {0.896} & {0.724} & {0.033} & {0.841} & {0.907} & {0.781} & 0.046 \\
    {OSFormer~\cite{pei2022osformer}} & {ECCV22} & {0.799} & {0.858} & {‡} & {0.073} & {0.811} & {0.881} & {‡} & {0.034} & {0.832} & {0.905} & {‡} & 0.049  \\
    {ZoomNet~\cite{pang2022zoom}} & {CVPR22} & {0.820} & {0.878} & {0.752} & {0.066} & {0.838} & {0.888} & {0.729} & {0.029} & {0.853} & {0.896} & {0.784} & 0.043 \\
    {FDNet~\cite{zhong2022detecting}} & {CVPR22} & {0.828} & {0.884} & {0.747} & {0.069} & {0.833} & {0.907} & {0.711} & {0.033} & {0.834} & {‡} & {0.750} & 0.052 \\
    {BGNet* ~\cite{sun2022boundary}} & {IJCAI22} & {0.812} & {0.870} & {0.723} & {0.080} & {0.827} & {0.894} & {0.707} & {0.033} & {0.853} & {0.903} & {0.785} & {0.045} \\
    {SINet-v2*~\cite{fan2022concealed}} & {TPAMI22} & {0.820} & {0.882} & {0.743} & {0.070} & {0.815} & {0.887} & {0.680} & {0.037} & {0.847} & {0.903} & {0.770} & 0.048 \\
    {FEDER~\cite{he2023camouflaged}} & {CVPR23} & {0.807} & {0.873} & {‡} & {0.069} & {0.823} & {0.900} & {‡} & {0.032} & {0.846} & {0.905} & {‡} & 0.045  \\
    {\textbf{\emph{FocusDiffuser-R}}} & \textbf{-} & {0.818} & {0.886} & {0.755} & {0.069} & {0.827} & {0.902} & {0.723} & {0.031} & {0.853} & {0.907} & {0.790} & 0.044  \\
    \midrule
    \multicolumn{14}{c}{Transformer based methods.} \\
    \midrule

    {VST~\cite{liu2021visual}} & {ICCV21} & {0.805} & {‡} & {0.780} & {0.069} & {0.810} & {‡} & {0.680} & {0.035} & {0.830} & {‡} & {0.740} & {0.053} \\
    
    {TPRNet~\cite{zhang2023tprnet}} & {TVCJ22} & {0.814} & {‡} & {0.781} & {0.076} & {0.829} & {‡} & {0.725} & {0.034} & {0.854} & {‡} & {0.790} & 0.047 \\
    {DTINet~\cite{liu2022boosting}} & {ICPR22} & {0.856} & {0.916} & {0.796} & {0.050} & {0.824} & {0.896} & {0.695} & {0.034} & {0.863} & {0.917} & {0.792} & 0.041  \\

    {ICON~\cite{zhuge2022salient}} & {TPAMI22} & {0.840} & {‡} & {0.769} & {0.058} & {0.818} & {‡} & {0.688} & {0.033} & {0.858} & {‡} & {0.782} & 0.041  \\
    
    {FPNet~\cite{cong2023frequency}} & {MM23} & {0.852} & {0.905} & {0.806} & {0.056} & {0.850} & {0.913} & {0.748} & {0.029} & {‡} & {‡} & {‡} & ‡  \\
    {OPNet~\cite{mei2023camouflaged}} & {IJCV23} & {0.858} & {0.915} & {0.817} & {0.050} & {0.857} & {0.919} & {0.767} & {0.026} & \underline{0.883} & {0.932} & {0.838} & 0.034  \\
    {HitNet~\cite{hu2023high}} & {AAAI23} & {0.844} & {0.902} & {0.801} & {0.057} & {\underline{0.868}} & {0.932} & {\underline{0.798}} & {\underline{0.024}} & {0.870} & {0.921} & {0.825} & 0.039  \\
    {FSPNet~\cite{huang2023feature}} & {CVPR23} & {0.856} & {0.899} & {0.799} & {0.050} & {0.851} & {0.895} & {0.735} & {0.026} & {0.879} & {0.915} & {0.816} & 0.035  \\
   
    {\textbf{\emph{FocusDiffuser-V}}} & \textbf{-} & \underline{0.869} & \underline{0.931} & \underline{0.842} & \underline{0.043} & 0.863  & \underline{0.934}  & 0.785  & \underline{0.024} & {0.882} & \underline{0.933} & \underline{0.840} & \underline{0.032} \\
    {\textbf{\emph{FocusDiffuser-P}}} & \textbf{-} & \textbf{0.881} & \textbf{0.939} & \textbf{0.851} & \textbf{0.042} & \textbf{0.875} & \textbf{0.939} & \textbf{0.809} & \textbf{0.020} & \textbf{0.891} & \textbf{0.940} & \textbf{0.854} & \textbf{0.029} \\
    \bottomrule
    \end{tabular}%
   }
  \label{tab:main_results}%
  \vspace{-1em}
\end{table}%

\begin{figure*}[pt]
   \centering
      \includegraphics[width=0.98\linewidth]{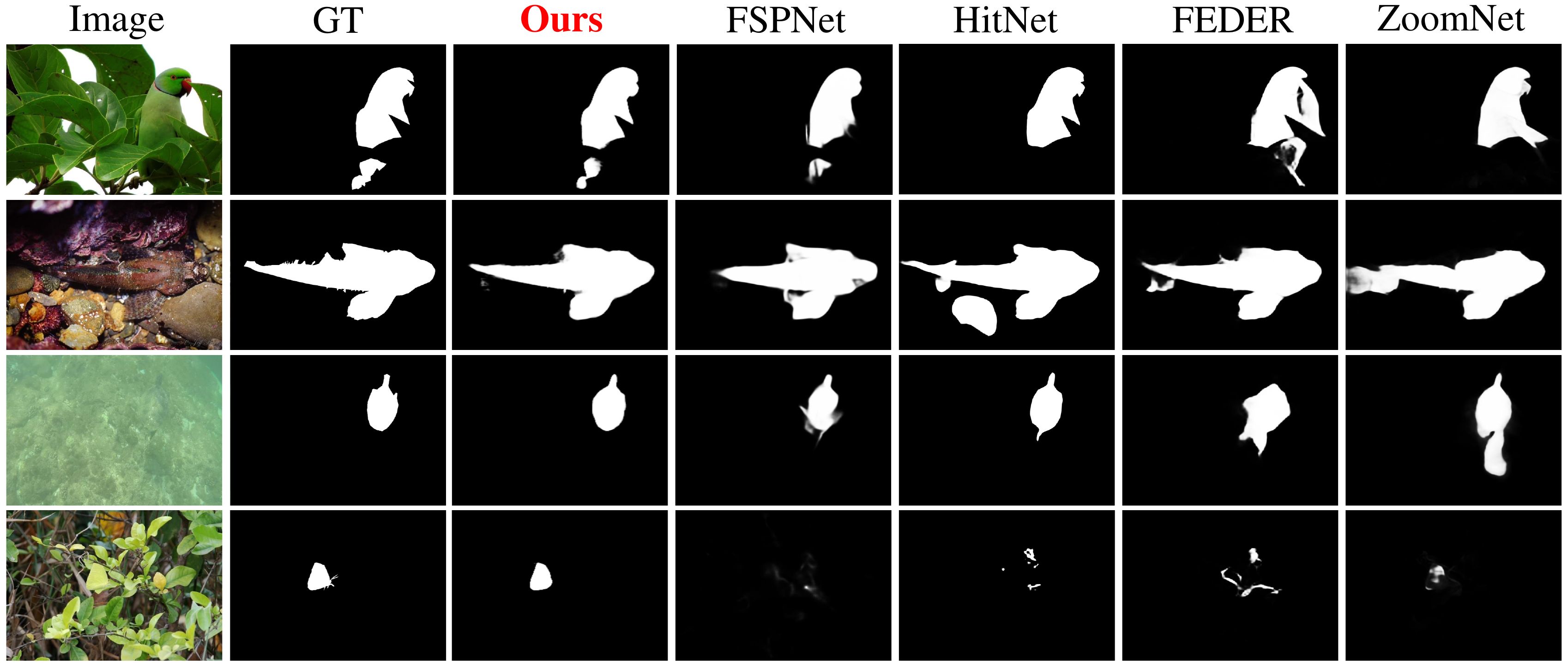}
    \vspace{-1em}
    \caption{{{\bf Qualitative comparisons} of predicted camouflaged maps with state-of-the-art models.} \label{fig:visualization}}.
    \vspace{-2.3em}
\end{figure*}

\begin{table}[pt]
    \begin{minipage}{0.46\textwidth}
    \caption{{\bf Ablation study} for the effectiveness of the Diffusion Model (DM).}
    \vspace{-1em}
          \resizebox{0.99\linewidth}{!}{
            \begin{tabular}{ccccccccccc}
            \toprule
            \makebox[0.08\textwidth][c]{\multirow{2}[1]{*}{\rotatebox[origin=c]{270}{PVT}}} &
            \makebox[0.08\textwidth][c]{\multirow{2}[1]{*}{\rotatebox[origin=c]{270}{Trans.}}} & 
            \makebox[0.08\textwidth][c]{\multirow{2}[1]{*}{\rotatebox[origin=c]{270}{DM}}} & 
            \multicolumn{4}{c}{COD10K(2026)} & \multicolumn{4}{c}{NC4K(4121)} \\
            \cmidrule{4-11} 
             &     &     & {$S_{\alpha}$ \boldmath{$\uparrow$}} & {$E_{\phi}$ \boldmath{$\uparrow$}} & {$F^{w}_{\beta}$ \boldmath{$\uparrow$}} & {$M$ \boldmath{$\downarrow$}} & {$S_{\alpha}$ \boldmath{$\uparrow$}} & {$E_{\phi}$ \boldmath{$\uparrow$}} & {$F^{w}_{\beta}$ \boldmath{$\uparrow$}} & {$M$ \boldmath{$\downarrow$}} \\
            \cmidrule{1-11}          
            \ding{52}&\ding{52}  &  & {0.815} & {0.865} & {0.673} & {0.044} & {0.840} & {0.884} & {0.740} & 0.050  \\
            \ding{52}&  &\ding{52}  & {0.853} & {0.914} & {0.787} & {0.033} & {0.871} & {0.918} & {0.826} & 0.038  \\
            \bottomrule 
            \label{tab:DM}
        \end{tabular}%
        }
    \end{minipage}
		\quad\quad
		\begin{minipage}{0.45\textwidth}
			\caption{{\bf Ablation study} for different components of \emph{FocusDiffuser}.}
        \vspace{-1em}
          \resizebox{0.95\linewidth}{!}{
            \begin{tabular}{ccccccccccc}
            \toprule
            \makebox[0.08\textwidth][c]{\multirow{2}[1]{*}{\rotatebox[origin=c]{270}{Base.}}} & 
            \makebox[0.08\textwidth][c]{\multirow{2}[1]{*}{\rotatebox[origin=c]{270}{BDLU}}} & 
            \makebox[0.08\textwidth][c]{\multirow{2}[1]{*}{\rotatebox[origin=c]{270}{CP}}} & \multicolumn{4}{c}{COD10K(2026)} & \multicolumn{4}{c}{NC4K(4121)} \\
            \cmidrule{4-11} 
             &            &    & {$S_{\alpha}$ \boldmath{$\uparrow$}} & {$E_{\phi}$ \boldmath{$\uparrow$}} & {$F^{w}_{\beta}$ \boldmath{$\uparrow$}} & {$M$ \boldmath{$\downarrow$}} & {$S_{\alpha}$ \boldmath{$\uparrow$}} & {$E_{\phi}$ \boldmath{$\uparrow$}} & {$F^{w}_{\beta}$ \boldmath{$\uparrow$}} & {$M$ \boldmath{$\downarrow$}} \\
            \cmidrule{1-11}          
            \ding{52}&  &           & {0.853} & {0.914} & {0.787} & {0.033} & {0.871} & {0.918} & {0.826} & 0.038  \\
            \ding{52}&\ding{52}&    & {0.868} & {0.923} & {0.798} & {0.027} & {0.880} & {0.924} & {0.837} & 0.036  \\
            \ding{52}& &\ding{52}   & {0.871} & {0.932} & {0.802} & {0.022} & {0.887} & {0.936} & {0.848} & 0.032  \\
            \ding{52}&\ding{52}&\ding{52}& \textbf{0.875} & \textbf{0.939} & \textbf{0.809} & \textbf{0.020} & \textbf{0.891} & \textbf{0.940} & \textbf{0.854} & \textbf{0.029} \\
            \bottomrule 
           \label{tab:equipment}
           \end{tabular}%
            }
		\end{minipage}
        \vspace{-2.5em}
	\end{table}
\subsubsection{Qualitative Analysis:} 
We benchmarked \emph{FocusDiffuser} against four top SOTAs, highlighting its excellence in Fig.~\ref{fig:visualization} with data from original studies. Challenges include blending objects with vague boundaries, subtle edges, and similar textures to the background. Existing models often underperform in these areas, whereas \emph{FocusDiffuser} produces maps with clear object outlines, improved edges, and accurate object delineation. This demonstrates \emph{FocusDiffuser}'s unmatched precision in detecting and enhancing camouflaged details for superior visual clarity.

\subsection{Ablation Study}

\subsubsection{Effectiveness of the Diffusion Model:} To highlight the Diffusion Model's (DM) exceptional capabilities, we compared it against a baseline model with a standard decoder (2 transformer layers from Segmenter~\cite{strudel2021segmenter}), keeping all settings unchanged. Results in Tab.~\ref{tab:DM} demonstrate the DM's superiority, showing marked improvements in $E_{\phi}$, $F^{w}_{\beta}$, and $M$. These metrics, essential for evaluating prediction mask quality, underscore the DM's advanced representational efficacy.

\subsubsection{Ablation for Boundary-Driven LookUp (BDLU):} 
\begin{figure*}[pt]
   \centering
      \includegraphics[width=0.98\linewidth]{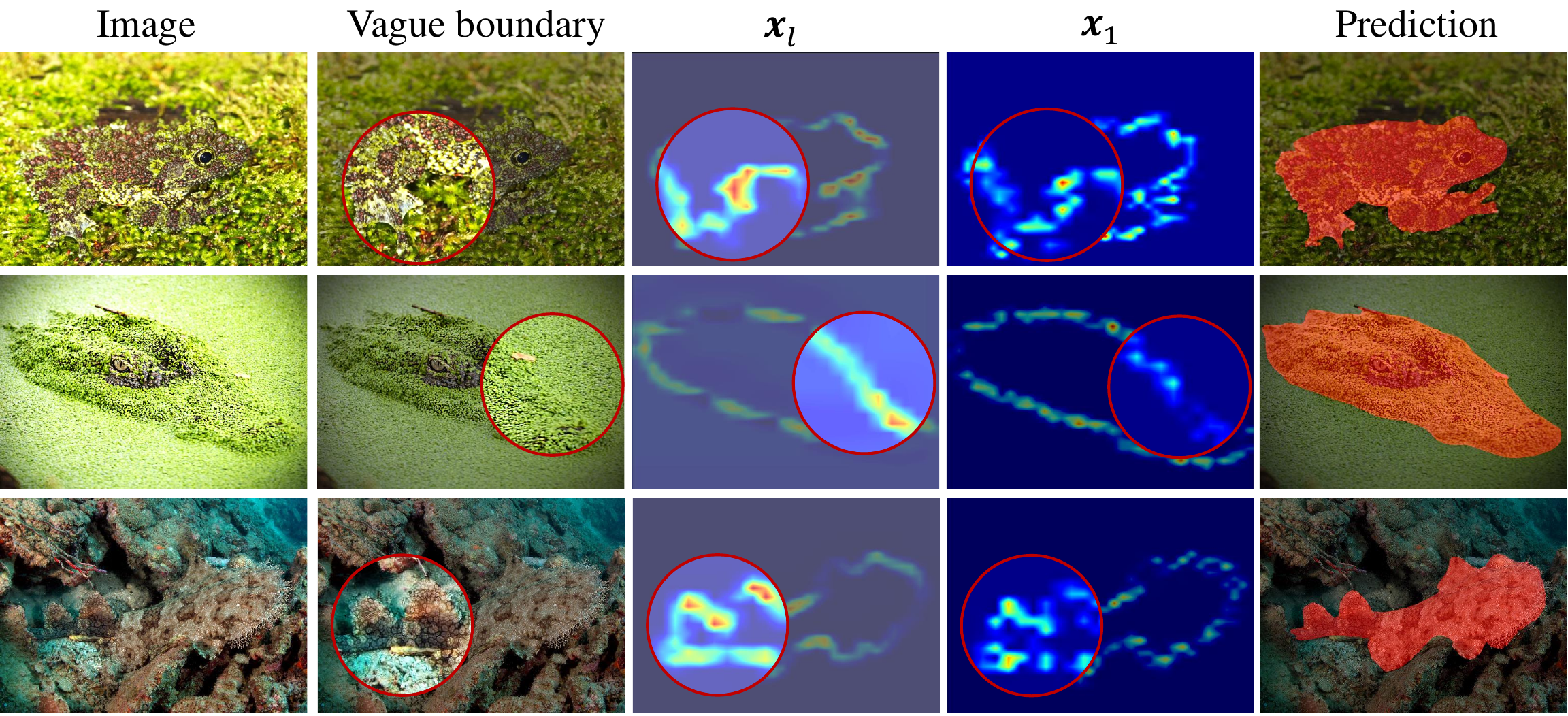}
    \vspace{-1em}
    \caption{{Through {\bf visualization} of the BDLU feature \(\boldsymbol{\mathit{x}}_l\), we observe that \emph{FocusDiffuser} specifically targets the most indistinct edges, where the texture of the object closely mimics that of the surrounding background, enabling it to delineate more precise boundaries. This heightened accuracy comes from the BDLU module's ability to identify local subtle differences, significantly boosting the model's performance in challenging situations.
    } \label{fig:ablation-BDLU}}.
    \vspace{-2.5em}
\end{figure*}

BDLU is crafted to enhance edge discrimination by harnessing local grid similarities. We visualized the feature representation \(\boldsymbol{x}_l\) to clarify BDLU's focus on areas where boundary distinction is most challenging, as illustrated in Fig.~\ref{fig:ablation-BDLU}. Quantitative evaluations, shown in Tab.~\ref{tab:equipment}, reveal a significant performance decline across all metrics without BDLU in \emph{FocusDiffuser}, underscoring its essential role. Further adjustments to BDLU configurations were explored to assess its impact on detecting camouflaged objects. Removing boundary selection and lookup operation, directly incorporating \(\boldsymbol{x}^{edge}_0\) into the denoising process resulted in a marked performance drop. Solely feeding \(\boldsymbol{x}_0^{edge}\) without boundary selection into LookUp module further degraded all metrics.

Alternatively, discarding the lookup operation and utilizing the boundary selection feature \(\boldsymbol{x}_1\) for the denoising process achieved the second-best performance, highlighting the edge features' significance in the lookup process. Visual comparisons between \(\boldsymbol{x}_l\) and \(\boldsymbol{x}_1\) (refer to Fig.~\ref{fig:ablation-BDLU}, columns 3 and 4) show the former's enhanced clarity around ambiguous boundaries. The lookup's efficacy is determined by the local grid radius, \(r\), with an optimal \(r\) being crucial for identifying similar features and clarifying indistinct boundaries. Optimum performance was observed with \(r\)=2. As \(r\) increases, it introduces extraneous features, diminishing the model's capability to discern local variations, as documented in Tab.~\ref{tab:BDLU-CP-detail}.

\subsubsection{Ablation for Cyclic Positioning (CP):} 
\begin{figure*}[pt]
   \centering
      \includegraphics[width=0.98\linewidth]{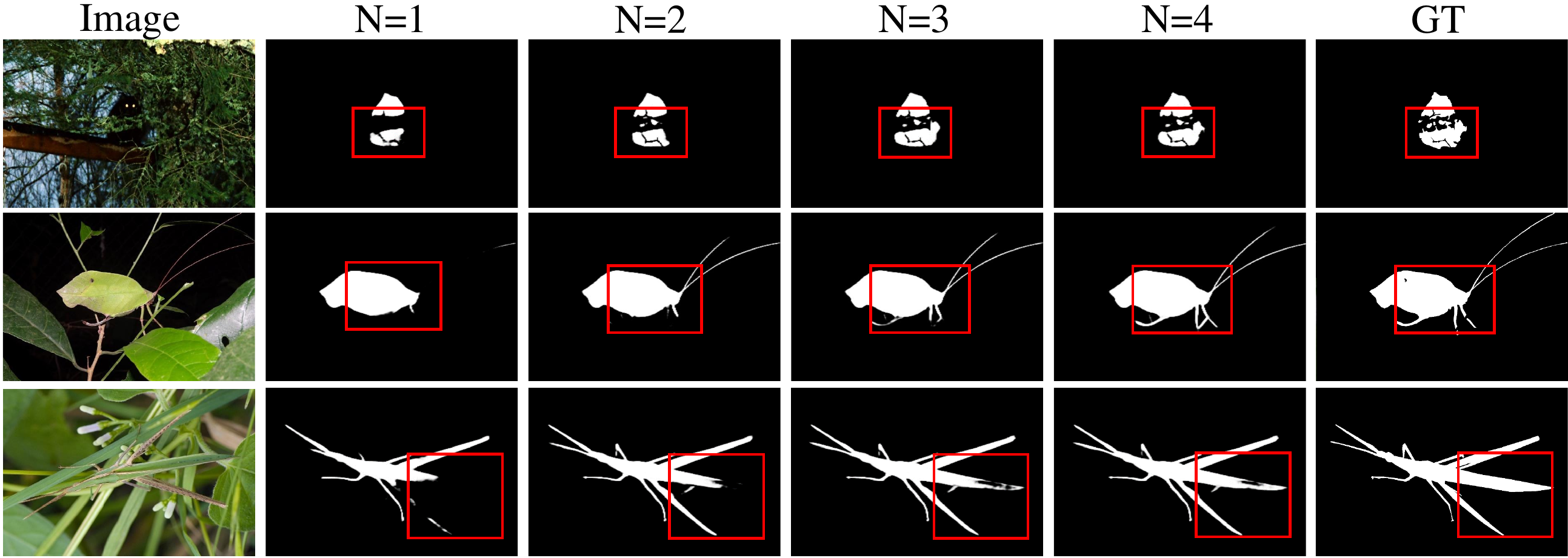}
    \vspace{-1em}
    \caption{{{\bf Illustration} of the enhancements applied to generative camouflaged maps through refinements by the cyclic positioning module. \emph{Zoom in for details}.} \label{fig:visual-cp}}.
    \vspace{-2.5em}
\end{figure*}

CP is leveraged to pinpoint camouflaged objects with the aid of a coarse map, integrating prior object location knowledge to extract targeted rectangular areas from the original image and combine these with edge details. This approach emphasizes the object, enriching texture detail visibility and ensuring prediction precision and granularity. Fig.~\ref{fig:visual-cp} depicts the CP mechanism, illustrating the iterative enhancement of prediction maps. Quantitative assessments in Tab.~\ref{tab:equipment} suggest that omitting CP in \emph{FocusDiffuser} leads to diminished outcomes. Moreover, three mask types, \emph{M}, are evaluated to gauge the effect of background suppression on camouflage detection (see Tab.~\ref{tab:BDLU-CP-detail}): `a' normalizes the predicted camouflaged map \(\mathbf{f}_0^{'}\) for \emph{M}, yet its subpar quality might exclude parts of camouflaged objects, yielding less optimal results than `b'. Contrarily, `b' uses \(\mathbf{f}_0^{'}\) as a basis for a bounding box that encases camouflaged objects, creating a proposal mask \emph{M} that better conserves semantic information and enhances detection robustness versus the coarse mask. `c', setting \emph{M} to 1 as a control, shows the least effectiveness without background exclusion.

\subsubsection{Ablation for Denoising Steps \(N\):}
We systematically investigate the influence of denoising steps \(N\) during the inference phase. Tab.~\ref{tab:denoising} displays the performance enhancement with increased \(N\) from 2 to 4 for scale factors \(b\) of 0.5 and 0.1. Nevertheless, we observed limited gain or even decline when \(N\) exceeds 4.

\subsubsection{Ablation Study for Scale Factor \(b\):}
\label{sec:scale-factor-b}
\(b\) is employed to regulate the signal-to-noise ratio of FD. As depicted in Fig.~\ref{fig:scale factor b}, initially setting \(b\) to 1 allows the noisy mask to remain distinguishable even at \emph{t}=600 timestamps, suggesting a high signal-to-noise ratio. Consequently, this may lead to an abundance of straightforward training examples for \emph{FocusDiffuser}, potentially resulting in a suboptimal model. In contrast, when \(b\) is reduced to 0.1, the noisy mask becomes challenging to observe at \emph{t}=600. Tab.~\ref{tab:denoising} outlines the changes in metrics as \(b\) decreases. Optimal performance for \emph{FocusDiffuser} is achieved with \(b\)=0.1.

\section{Conclusion}
\label{sec:conclusion}
We introduce \emph{FocusDiffuser}, a conditional diffusion model featuring a Bound-ary-Driven LookUp (BDLU) and a Cyclic Positioning (CP) module, engineered to enhance the detection of camouflaged objects against their environments. The BDLU module aggregates similarity values within a local grid to refine feature representations, emphasizing areas where camouflaged objects blend with their surroundings. Concurrently, CP identifies and isolates camouflaged objects, creating a focus region matrix that separates them from the background. Our extensive experiments demonstrate these modules synergistically improve the identification and clarity of camouflaged objects in COD tasks. Furthermore, we posit \emph{FocusDiffuser}'s methodology offers potential advancements in other computer vision areas requiring precise, high-quality predictions, such as high-resolution segmentation.


\section*{Acknowledgements}

This research was supported by the National Natural Science Foundation of China (62203089, 62303092, 62103084); the National Key Research and Development Program of China (No. 2022YFB2503004); and the Sichuan Science and Technology Program (2022NSFSC0890, 2022NSFSC0865, 2021YFS0383, 2023YF G0024, 2022YFS0570, 2023YFS0213).


%
%
\bibliographystyle{splncs04}
\bibliography{main}
\end{document}